# NEURAL NETWORK CAPACITY FOR MULTILEVEL INPUTS

Matt Stowe and Subhash Kak
.

**Abstract:** This paper examines the memory capacity of generalized neural networks. Hopfield networks trained with a variety of learning techniques are investigated for their capacity both for binary and non-binary alphabets. It is shown that the capacity can be much increased when multilevel inputs are used. New learning strategies are proposed to increase Hopfield network capacity, and the scalability of these methods is also examined in respect to size of the network. The ability to recall entire patterns from stimulation of a single neuron is examined for the increased capacity networks.

TABLE OF CONTENTS

Chapter Page

## I INTRODUCTION

This paper is devoted to the question of memory capacity of artificial neural networks. These networks are of two types: feedforward and feedback, and they typically use binary neurons. The basic model of learning used is Hebbian [1] in which neurons firing in the same manner get their connection strengthened and neurons firing in the opposite way get their interconnection weakened. The Hopfield (or feedback) neural network model [2],[3], which may be viewed as a generalization of the idea of storage in terms of eigenvectors for a matrix, is a model for storage. It is not a model for recall by index. In this model we can only check if a given memory is stored, but since the memory is not localized it cannot be recalled by index.

Memories are viewed as stored in patterns of activity [4]-[6] and they may be indexed [7]-[11]. This seems to be particularly relevant in the ability of individuals to recall a tune from a note, and the fact that we can recognize objects and individuals even when seen in strikingly different situations. This indexing can be aided by cures from different memory sub-systems. Another view is that information stored in specific locations affects how new memories are stored as in some recent experimental findings [12],[13]. The memory capacity of the neurons for assumed patterns of activity is roughly equal to the number of neurons if they are binary [14] and, as expected, higher if they are non-binary [15]. But the increase in the memory capacity upon the use of non-binary inputs was not fully investigated and this is what will be done in this paper.

Very broadly, there are two categories of memory: short-term memory (or working memory) and long-term memory. Long-term memory is the brain's system for storing, managing, and retrieving information whereas short-term memory or working memory does its functions in the mind before either being dismissed or transferred to long-term memory. If artificial neural networks are to model short-term memory, they should have the capacity to generalize very quickly; this is indeed possible in certain models [4]-[6]. Popular learning algorithms such as the backpropagation algorithm take much longer to store memories and, therefore, they could serve as models of long-term memory. But the processing strategies in the brain are bound to be much more complex than any simple artificial model where all neurons are on the same hierarchical plane. Indeed, there is evidence that certain neurons are activated by specific scenes or letter strings [16] and that neural cliques operate at different levels [17]-[21].

Long-term memories are much more complex than short-term ones. Different types of information (such as procedures, life experiences, and language) are stored with separate memory systems. Explicit memory, or declarative memory, is a type of long-term memory, which requires conscious thought. It's this that is normally meant by memory when judgment about somebody's memory being good or bad is made. Explicit memory is often associative in a manner that links different categories together [22]-[24].

Quantum models of memory have been proposed. One proposal is in terms of virtual bosons associated with the physiological structures of the brain in which long term memory is related to the ground state and short-term memory to the meta-stable excited states [25]-[30]. Quantum models are fundamentally two-tiered. They assume a deeper quantum memory that is tied to a neural-network based conscious memory [31]-[33]. Other models assume that humans encode not the full correlational structure of the input, but rather higher level representations of it [34]-[36]. It should also be noted that cognitive processing has paradoxical aspects some which are related to limitations of logical analysis [37]-[40].

The brain is composed of several modules, each of which is essentially an autonomous neural network. Thus the visual network responds to visual stimulation and also during visual imagery, which is when one sees with the mind's eye. Likewise, the motor network produces movement and it is active during imagined movements. Despite the modular nature of the brain, a part of it which is located for most people in the left hemisphere monitors the modules and interprets their individual actions in order to create a unified idea of the self. In other words, there is a higher integrative or interpretive module that synthesizes the actions of the lower modules [41]-[44].

The Hopfield network comprises of neurons where each neuron has a weighted connection to every other neuron, but neurons do not feed into themselves. A pattern is stored in such a network by strengthening the connections between neurons. Neurons that are activated together have their connection weights to each other increased, and neurons that are not activated together have those weights decreased (Hebbian learning). Hopfield networks will converge to a pattern, but it might not be the desired pattern. Patterns recognized by a Hopfield network converge to themselves when activated. If a pattern that is close to a stored pattern is activated, the stored pattern will be converged to. This behavior is useful for vocal and facial recognition programs. An example of an artificial neuron is shown in figure 1.1.

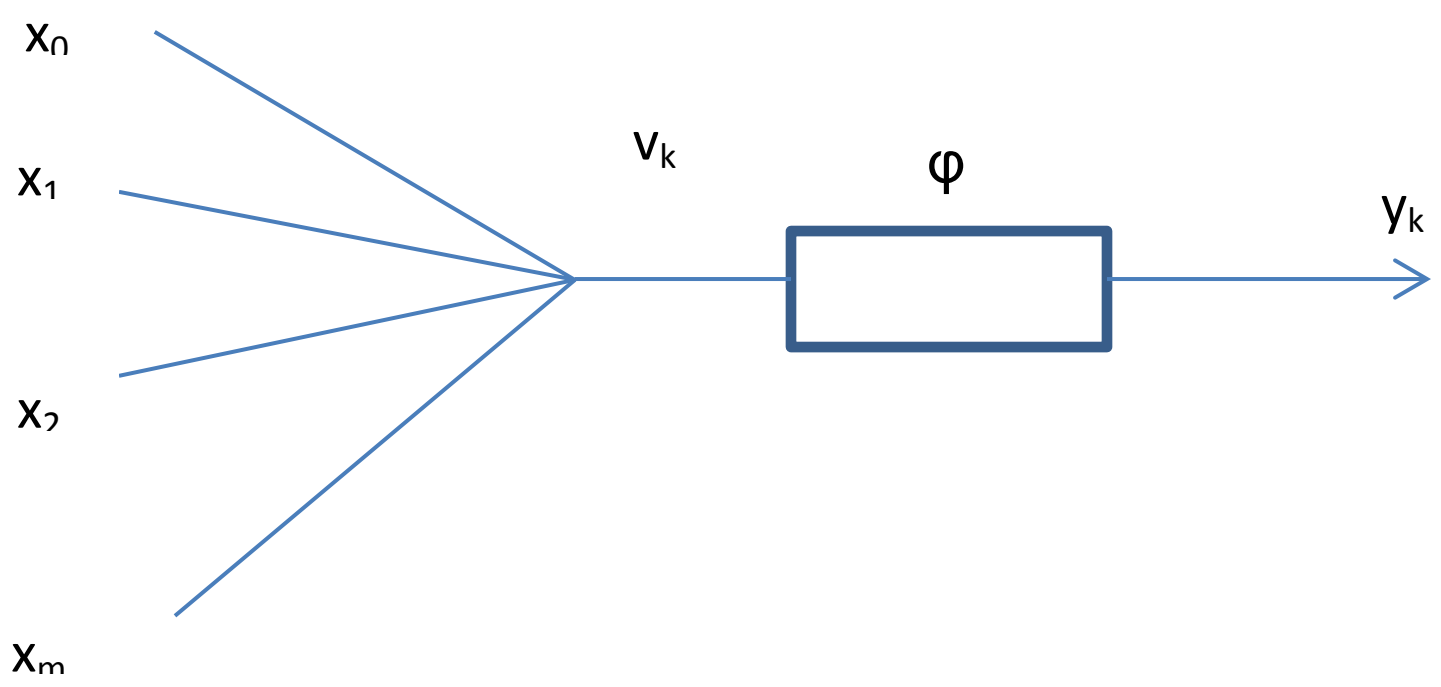

Fig. 1.1 Model of an artificial neuron

There are several basic properties shared by different models of artificial neurons. In figure 1.1, there are $m$+1 inputs with $x_0$ through $x_m$ inputs and $w_0$ through $w_m$ weights. Only $m$ inputs are available in figure 1.1 because $x_0$ is set to +1. Equation 1.1 defines $y_k$.

$$y_k = \varphi\left(\sum_{j=0}^{m} w_{kj} x_j\right) \tag{1.1}$$

The inputs of an artificial neuron are analogous to the dendrites of a biological neuron. The summation function represents the soma of a biological neuron. The outputs represent axons. Each of these structures is significantly simplified from its biological version. In particular, the axon has a continuous range of output values. Dendrites also use a large number of their numbers to achieve a large range of input values. Linear thresholding is applied in artificial neurons in order to better approximate these behaviors.

There are several types of transfer functions used by artificial neurons. The transfer function is represented by $\varphi$ in figure 1.1 and equation 1.1. The step function is used in many models. If a threshold is met, then the signal is sent. Otherwise, the signal is inhibited. This paper uses the step function in its basic form and also uses linear thresholding to widen the range of possible outputs. Linear combination adds the inputs to a bias term and returns this as the output. Sigmoid functions can also be used. A sigmoid function is bounded by two horizontal asymptotes and is shaped like the letter ‘S’ when plotted on a graph. A useful property of sigmoid functions is their derivative. Sigmoid functions always have a positive derivative. This allows for easy computation of changes between updates of an artificial neural network.

Artificial neural networks and artificial neurons model a simplified version of biological neural networks, or neural pathways. The neurons within a biological neural network are very complicated and difficult to model artificially. The connections between neurons are chemical synapses, electrical gap junctions, and other more advanced connections instead of the simple weight values used by Hopfield networks. Activation of a biological neuron can cause an action potential or spike which will cause the neuron to signal other neurons. The formation and recall of human memory is still unknown. Theories have been formulated involving complicated dynamics in biological neural networks and the complex systems within them, but no concrete explanation exists.

The biological neuron consists of dendrites, a soma, and an axon. The soma is the cell body of the neuron, which dendrites and the axon connect to. Dendrites have a lot of branches, and each branch is usually thinner than its trunk. Dendrites usually extend a few hundred micrometers from the soma. An axon can extend much farther, and does not get thinner. Although there is only one axon, that axon can branch hundreds of times. A soma can have multiple dendrites but only one axon. Typically, the axon of a neuron connects to a dendrite or soma of another neuron. However, biological neurons are extremely varied, and some neurons do not have dendrites or an axon. There are also neurons that connect dendrites to other dendrites, axons to other axons, and dendrites to axons. Neurons signal other neurons by firing synapses from the sending neuron's axon to the receiving neurons' dendrites and/or somas. The synapses fired can either excite or inhibit the receiving neuron. If a neuron receives enough excitatory signals within a short amount of time, then an action potential is generated. An action potential is a pulse that fires along the axon and activates other neurons. Action potentials facilitate interneural communication by jumping the gap between neurons to send signals.

This paper will investigate the capacity of generalized feedback neural networks, which most likely model biological neural networks better than artificial networks composed of binary neurons.

## II MODELS

A biological neuron model attempts to accurately portray neurons and their interactions instead of attempting to achieve computational effectiveness. Biological neuron models are also known as spiking neuron models. Instead of dealing with abstract terms, actual physical components are used. Instead of simple numeric patterns, input to a neuron is often represented by an ion current which passes through the cell membrane. Input is triggered by neurotransmitters which activate ion channels in the cell. This activation sequence is described by the function $I(t)$ where $t$ is time.

The cell is surrounded by its membrane, which has a concentration of charged ions on either side which determines $C_m$ which is the cell's ability to store a charge, or the cell's capacitance. The output of the cell is a change in voltage. This change of voltage can result in a voltage spike known as an action potential. $V_m$ is the quantity of the voltage spike. Several biological neuron models have been proposed. All of the proposed biological neuron models have drawbacks however. Cells differ from each other in ways which must be accounted for, and temperatures are higher than experimental data. Many problems involving temperatures and non-uniformity are still unsolved, and new models are constantly being designed to better model physical neurons.

One of the simplest and earliest biological neuron models is merely the derivative of the law of capacitance $Q = CV$. A French neuroscientist by the name of Louis Lapicque proposed this method in 1907. His model is called integrate-and-fire and is shown by equation 2.1[45].

$$I(t) = C_m \frac{dV_m}{dt} \tag{2.1}$$

This model takes an input current and increases the membrane voltage over time. Once the voltage hits a constant threshold $V_{\mathrm{th}}$, an action potential is generated. The voltage is then reset to its resting potential before continuing to apply input current. This means that the action spike can happen repeatedly until the input current is completely utilized. This is not a very accurate depiction of an actual biological neuron so an improvement was made in the form of a refractory period $t_{\mathrm{ref}}$ which prevents the action potential from being generated. This is more like a biological neuron which requires a period of rest before generating another action potential. With this addition, the firing frequency can be shown as a function described by equation 2.2 [45].

$$f(I) = \frac{I}{C_m V_{th} + t_{ref} I} \tag{2.2}$$

Although this adjustment is an improvement, there is still at least one problem with the model. If an input current is not strong enough to trigger an action potential, then the partial triggering is retained indefinitely. In actual biological neurons, the charge dissipates over time. To account for this, the leaky integrate-and-fire model subtracts some charge based on time. This model is described by equation 2.3.

$$I(t) - \frac{V_m(t)}{R_m} = C_m \frac{dV_m(t)}{dt} \tag{2.3}$$

In this model, $R_m$ is the membrane resistance. This allows any charge that does not result in an action potential to dwindle over time. The resulting firing frequency is shown by equation 2.4.

$$f(I) = \begin{cases} 0, & I \leq I_{th} \\ \left[t_{ref} - R_m C_m \log\left(1 - \frac{V_{th}}{I R_m}\right)\right]^{-1}, & I \geq I_{th} \end{cases} \tag{2.4}$$

Nicolas Fourcaud-Trocmé, David Hansel, Carl van Vreeswijk and Nicolas Brunel made a simple modification to the integrate-and-fire model with the exponential integrate-and-fire model shown in equation 2.5[46].

$$\frac{dX}{dt} = \Delta_T exp\left(\frac{X - X_T}{\Delta_T}\right) \tag{2.5}$$

In this model, $X$ is the membrane potential, $X_T$ is the membrane potential threshold, and $\Delta_T$ is the sharpness of action potential initiation.

The integrate-and-fire models have been widely used because of their simplicity, but more accurate models also exist. One of the most successful models is the Hodgkin-Huxley model. This model has multiple currents instead of just one, as represented by equation 2.6 [47].

$$C_m \frac{dV(t)}{dt} = -\sum_i I_i(t, V) \tag{2.6}$$

Current is represented by $I_i$, and a single current is modeled by equation 2.7 [47].

$$I(t, V) = g(t, V) \times (V - V_{eq}) \tag{2.7}$$

The conductance is represented by $g$ and contains activation fraction $m$ and inactivation fraction $h$. The Hodgkin-Huxley model is very complex, and attempts to model every aspect of a biological neuron. Due to its complexity, several simpler models have been proposed. One simplification is the FitzHugh-Nagumo or FHN model shown in equation 2.8 [48].

$$\frac{dV}{dt} = V - V^3 - w + I_{ext} \tag{2.8}$$

Although vastly simplified, this model is close enough to Hodgkin-Huxley to be useful. The general gate voltage is represented by $w$ [48].

Another model based on the Hudgkin-Huxley is Morris-Lecar shown in equation 2.9[49].

$$C \frac{dV}{dt} = I - g_L(V - V_L) - g_{Ca} M_{ss}(V - V_{Ca}) - g_K N(V - V_k) \tag{2.9}$$

This model focuses on calcium and potassium channels, represented by $g_{Ca}$ and $g_K$ respectively. The giant barnacle muscle fiber exhibits a variety of oscillatory behaviors with these two channels, and those behaviors are the target of the Morris-Lecar model [49].

The Hindmarsh-Rose model [50] is based on the FitzHugh-Nagumo model. It consists of 3 equations 2.10, 2.11, and 2.12.

$$\frac{dx}{dt} = y + ax^2 - x^3 - z + I \tag{2.10}$$

$$\frac{dy}{dt} = 1 - bx^2 - y \tag{2.11}$$

$$\frac{dz}{dt} = r(s(x - x_R) - z) \tag{2.12}$$

The Hindmarsh-Rose model is still relatively simple compared to the Hodgkin-Huxley, but it also allows for a wide range of dynamic behaviors.

There are also biological neurons models for cell structure that is different the standard idea. However, in any model there will be some generalization and abstraction from the actual biological neuron. Models are still being created in attempts to more accurately capture the structures and behaviors of various biological neurons.

In this paper, artificial neurons are used. These are much simpler than their biological counterparts, and are targeted at the communication between neurons rather than the inner workings of the neurons themselves. Hopfield networks of artificial neurons are examined in this paper. Hopfield networks are represented by a square matrix whose size is the number of neurons in the network. Each row in the matrix represents a neuron, and each column represents a weight for the neuron. Since each neuron has no weight to itself, the diagonal of this matrix is 0. Initially, every neuron has a connection weight of zero to every other neuron.

Patterns are represented by a vector with a length equal to the number of neurons in the network. To store a pattern, various learning rules can be used. Hebbian learning is used for this paper, and is represented by equation 2.13.

$$T_{ij} = \sum_{s=0}^{m} V_i^s V_j^s \quad i \neq j \tag{2.13}$$

$T$ is the matrix of size $m$ by $m$, and $V$ is a neuron within the matrix. Hopfield networks trained with Hebbian learning have a capacity of .15$N$, where $N$ is the number of neurons. A network trained with Hebbian learning will also be symmetrical. If the pattern is not stored, delta learning can be used as shown in equation 2.14.

$$\Delta T_{ij} = c(V_i^s - V_i)V_j^s \tag{2.14}$$

$c$ is a small learning constant which is 1 for this paper. Delta learning adjusts the weights of neurons in the matrix toward the desired values and significantly increases the number of memories that can be stored in a Hopfield network [2]. Classic Hopfield networks only deal with binary patterns which either have a value of +1 or -1. Values greater than 0 are mapped to +1, and values less than or equal to 0 are mapped to -1. This mapping is represented by equation 2.15.

$$y = \begin{cases} 1 & x > 0 \\ -1 & x \leq 0 \end{cases} \tag{2.15}$$

*x* is the value to be mapped and *y* is the result of the mapping. The use of binary thresholds for classic Hopfield networks has come into question because biological networks have a continuum of values, and are probably not restricted to two distinct values [44]. Non-binary networks have been shown to have superior information storage capability [3]. A generalization of equation 2.15 allows linear thresholding for non-binary networks as shown in equation 2.16.

$$y = \begin{cases} r_0 & x > t_0 \\ r_1 & t_0 \geq x > t_1 \\ \vdots & \vdots \\ r_{n-2} & t_{n-3} \geq x > t_{n-2} \\ r_{n-1} & t_{n-2} \geq x \end{cases} \tag{2.16}$$

*t* is a list of thresholds, and *r* is a list of values that thresholding can result in. Prados and Kak explored the increased capacity gained when *n*=4. Various values for *t* were tested across various values of *r*. It was shown that using delta learning and non-binary networks provided a much higher capacity than binary networks without delta learning [3]. Binary networks with 7 neurons using delta learning could store 6 patterns 36% of the time. Table 1.1 shows the results of the paper when thresholding was applied.

| | | Number of Patterns Stored | | | | | |
|---|---|---|---|---|---|---|---|
| *r* | *t* | 1 | 2 | 3 | 4 | 5 | 6 |
| {-1 1} | {0} | 100 | 97 | 86 | 76 | 57 | 36 |
| {-2 -1 1 2} | {-24 0 24} | 100 | 100 | 100 | 97 | 88 | 50 |
| {-3 -1 1 3} | {-108 0 108} | 100 | 100 | 100 | 95 | 81 | 37 |
| {-4 -1 1 4} | {-96 0 96} | 100 | 100 | 100 | 96 | 85 | 47 |

Table 1.1

It is clear that linear thresholding improves the capacity of neural networks trained with delta learning. Not only is the number of patterns stored greater than binary networks, but each pattern stores more information because instead of two states of information being represented by a neuron, multiple states are represented in each neuron, resulting in an exponential growth in the capacity of neural networks. One pattern stored in a traditional binary Hopfield network with 8 neurons is $2^8$ possible values, or one byte of information. With thresholded values, the capacity increases exponentially. One pattern stored in a quaternary Hopfield network with 8 neurons is $4^8$ or 65536 possible values, or two bytes of information. The general formula using equation 2.16 is $n^N$ possible values, and $\log_2(n^N)$ bits of information.

## III METHODOLOGY

Several patterns were randomly generated to be stored in a Hopfield network. These patterns are at least 1 hamming distance apart from each other. This means that each pattern to be stored differs from every other pattern in at least 2 neurons. This is done because Hopfield networks as presented in this paper are incapable of storing patterns that only differ by 1 neuron since such patterns will not be associated with unique minima. This concept is illustrated in figure 3.1.

The red points along the line represent stable patterns in a Hopfield network. These stable patterns are local minima in an attraction basin. During training of the Hopfield network using Hebbian learning and the delta rule, the state of the network moves along the line to the desired local minima until the pattern is stored. Because of this, the black points cannot be stored because they are not stable patterns. However, the pattern might not store because the state of the network could jump over the local minima continuously.

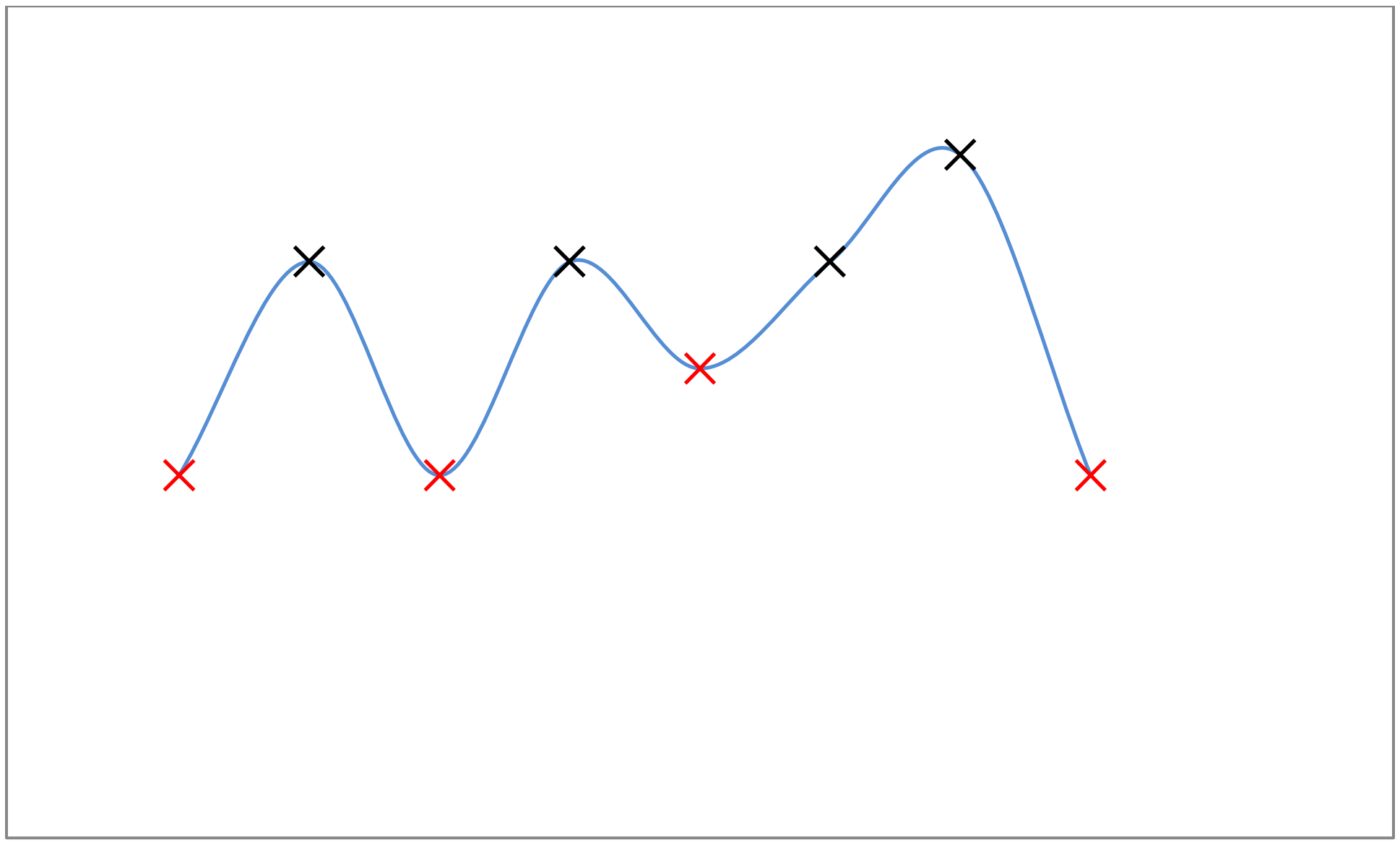

Fig. 3.1 Sample energy minima of the Hopfield network.

First, a pattern is stored with Hebbian learning. If the pattern fails to store, delta learning is repeatedly applied to the pattern until a maximum limit has been reached or the pattern is stored. Then the entire group of memories has delta learning applied again until a maximum limit is reached or all memories are stored. If the memories haven't been stored, the order of memories is changed and all memories are again treated to delta learning. A limit must be set on how many times to repeat each individual pattern, each group of patterns, and each ordering of patterns because it is possible that application of the delta rule will result in an infinitely repeating set of states, none of which store all the desired patterns. Various thresholds were tested, each with 100 sets of patterns.

In order to determine how much the patterns themselves affect storage capability, variable thresholds were tested. Testing variable thresholds starts out the same way as above, but if the patterns fail to store the thresholds are adjusted, and the entire storage procedure was attempted again. This continues until the patterns store or until a maximum number of threshold adjustments has been reached.

Storing single patterns was tested to determine the maximum amount of information capacity in bits that a 7 by 7 network can achieve. Since there was no group of patterns, delta learning was simply repeatedly applied to a single pattern in attempts to store it.

Recalling entire patterns from the stimulation of a single neuron was also attempted with thresholding. Each neuron was set to each possible value, and then the pattern was recalled using the recall method presented by Kak [37].

The scalability of the delta learning applications and various thresholds were also tested as far as capacity with respect to size of the network. Tests in varying thresholded networks were carried out for 10 by 10, 15 by 15, and 20 by 20 networks.

## IV. EXPERIMENTS RELATING CAPACITY WITH THRESHOLD LEVELS

Applying delta learning in different orders on the memories has shown superior storage capacity for networks. For the following graphs, the legend on the right represents the thresholds used, the vertical axis is the percentage successfully stored, and the horizontal axis is the number of patterns stored. In each graph, 6 patterns attempted to store in the same network.

Thresholding values to {-2 -1 1 2} gave the following results:

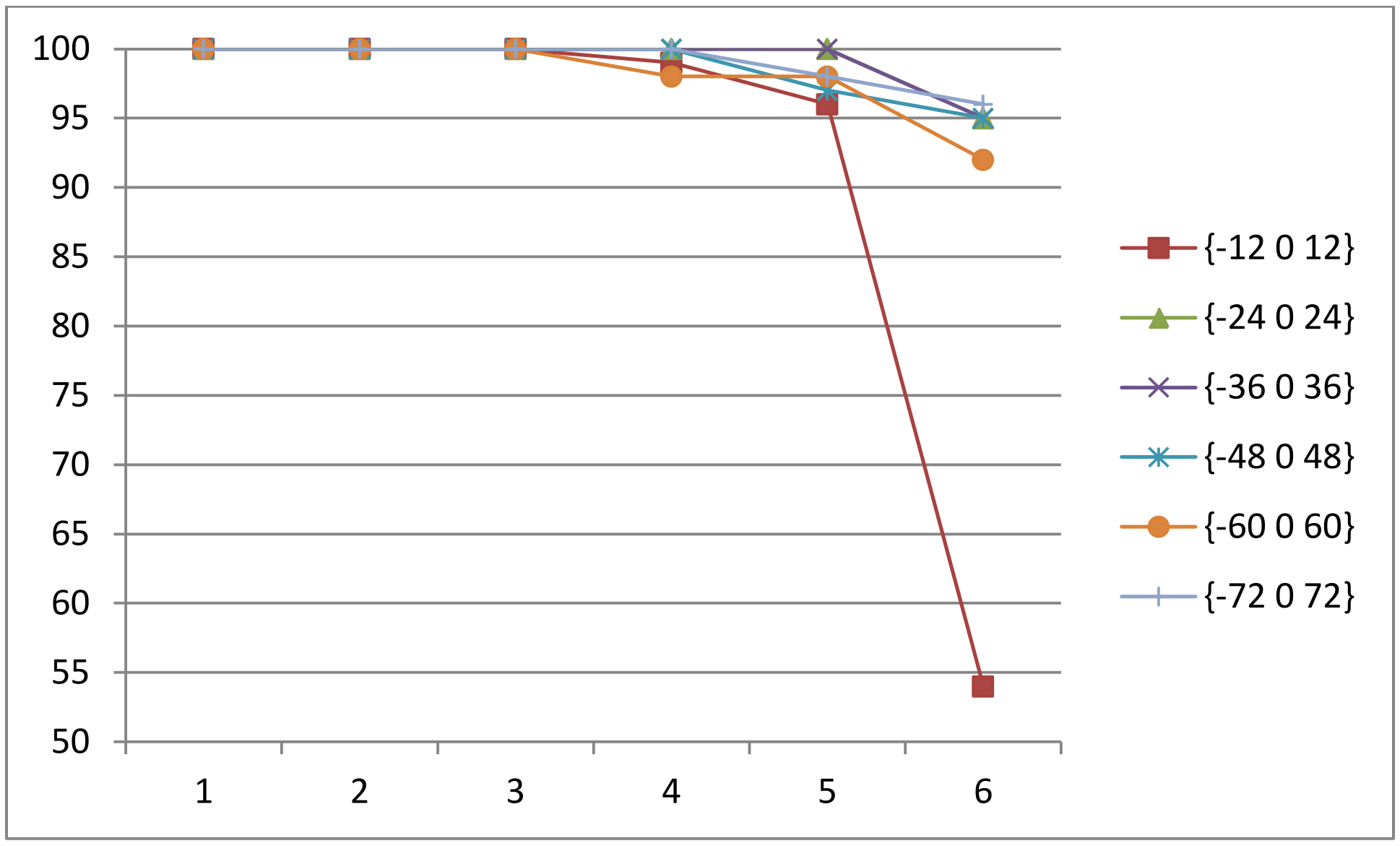

Fig. 4.1 Storing 6 patterns thresholding to {-2, -1, 1, 2}.

Thresholding values to {-3 -1 1 3} gave the following results:

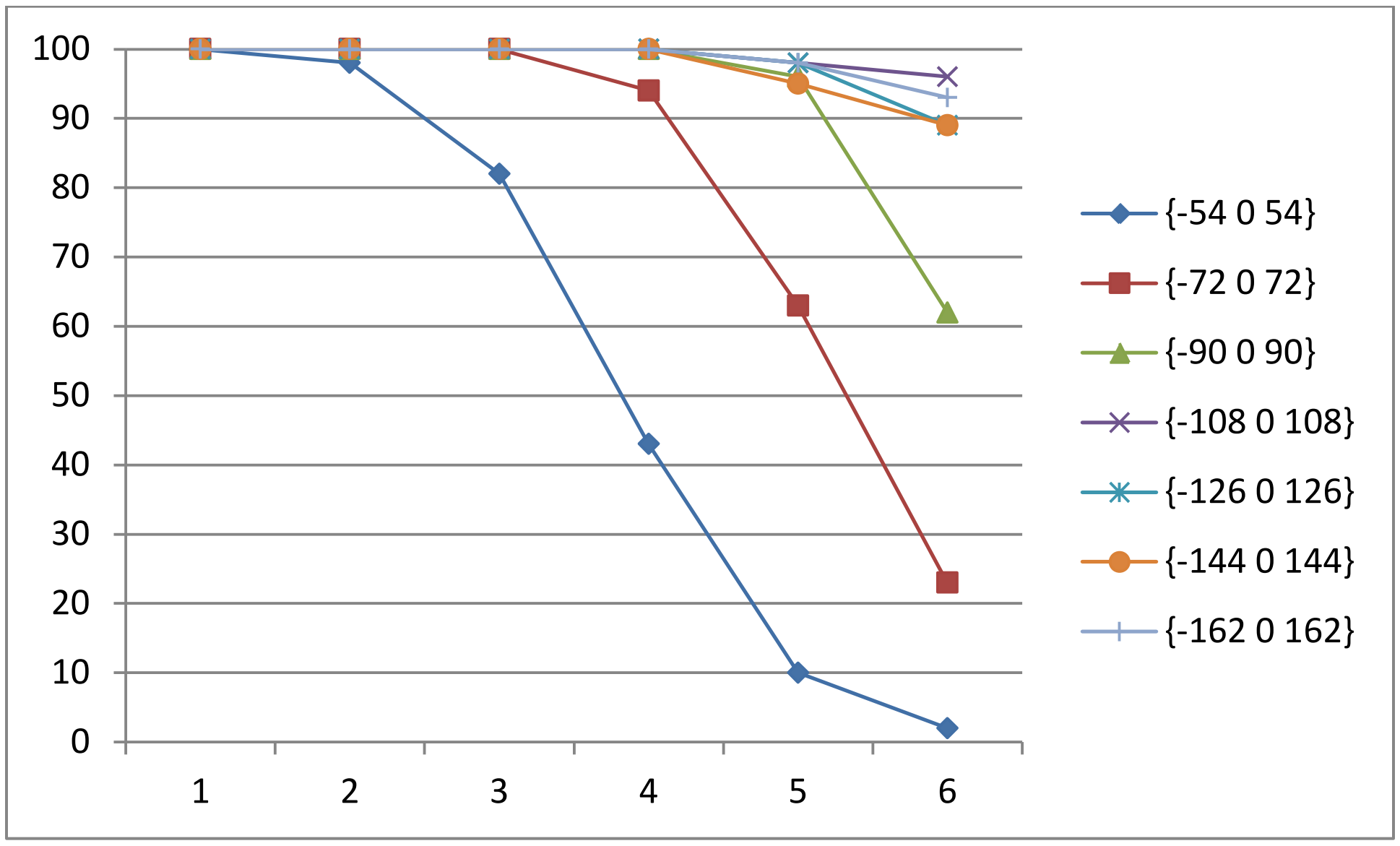

Fig. 4.2 Storing 6 patterns thresholding to {-3, -1, 1, 3}.

Thresholding values to {-4 -1 1 4} gave the following results

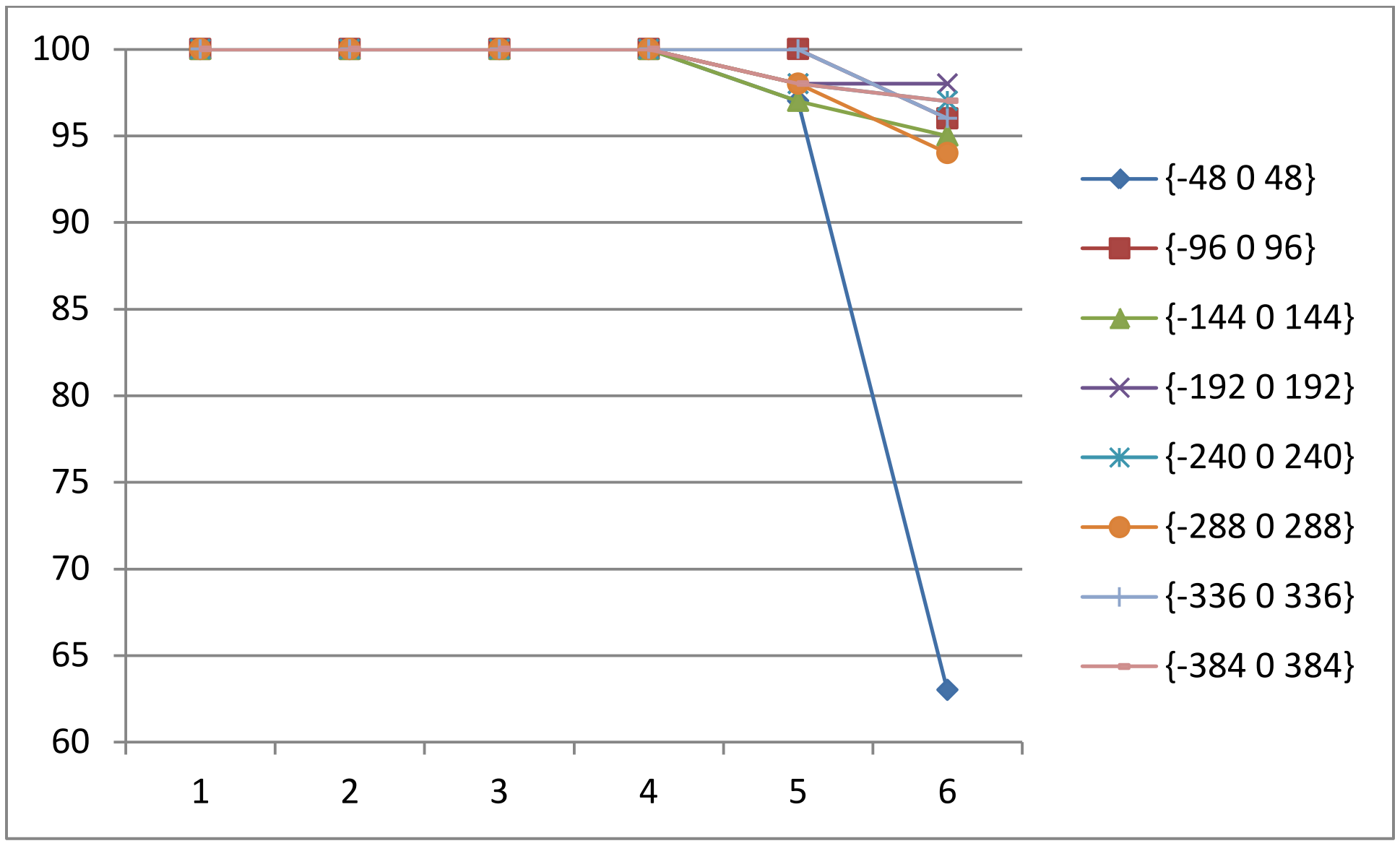

Fig. 4.3 Storing 6 patterns thresholding to {-4, -1, 1, 4}.

Attempts at storing larger numbers of patterns also met with success, though in smaller percentages. In rare cases, it is possible to store $2N$ distinct patterns. Testing 10 permutations of order of memories, and 100 attempts at storing 15 patterns, and 10 attempts at storing each individual pattern, the following results were achieved.

Thresholding values to {-4 -1 1 4} allowed for one set of 14 patterns to be stored:

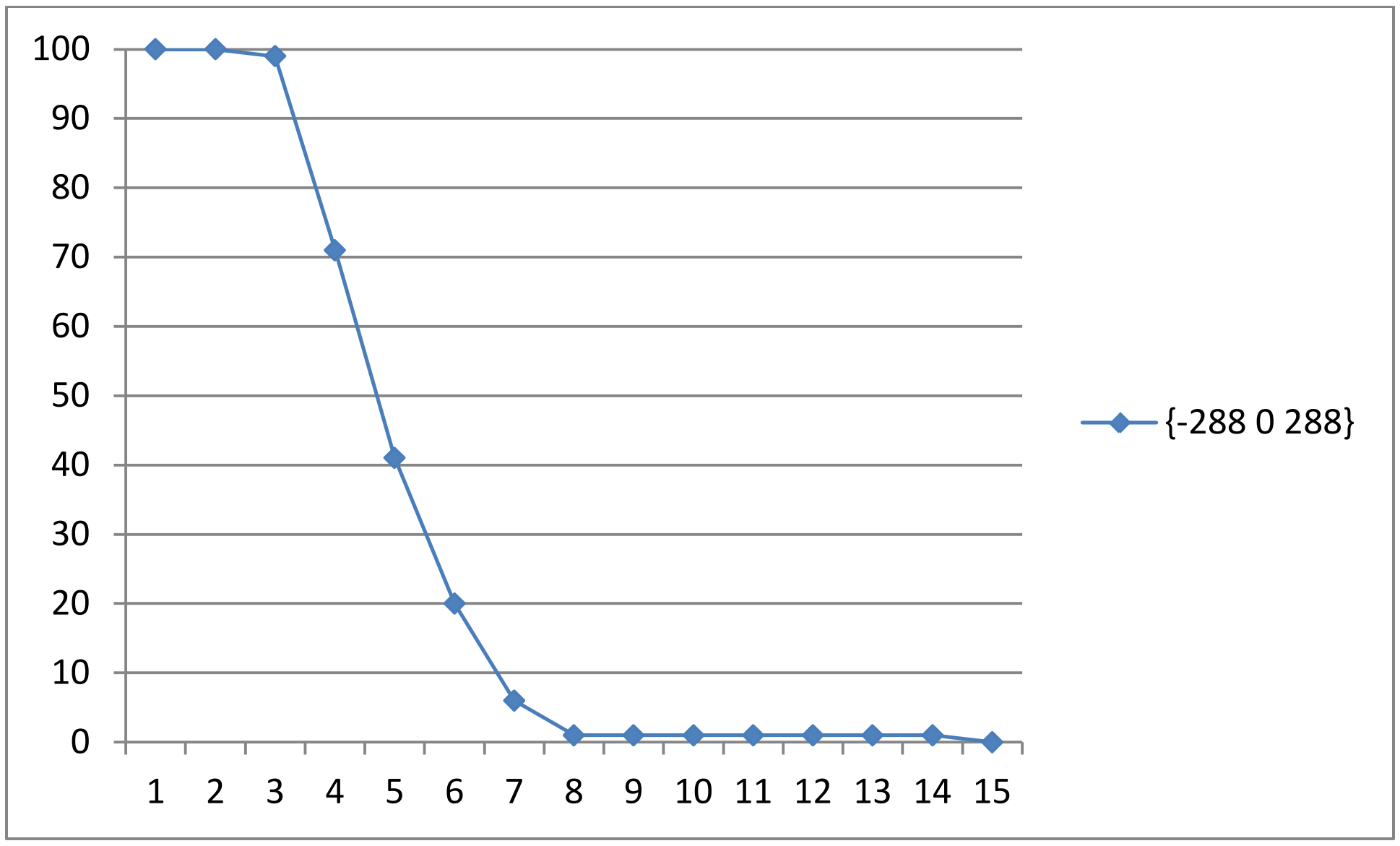

Fig. 4.4 Storing 14 patterns thresholding to {-4, -1, 1, 4}.

The fifteen patterns that were attempted to be stored were:

| | | | | | | | | | |
|---|---|---|---|---|---|---|---|---|---|
| {1 | -1 | -1 | 4 | 1 | -1 | -4 | -1 | 1 | 1 } |
| {1 | -1 | -4 | -1 | -4 | 1 | 1 | 4 | -4 | -4} |
| {1 | 1 | 4 | 4 | 4 | 4 | 1 | -4 | -4 | 4 } |
| {1 | -1 | 4 | 4 | -1 | -1 | -1 | -4 | 4 | -1} |
| {-4 | -4 | 4 | -4 | 4 | -4 | 4 | -1 | -4 | -4} |
| {1 | -1 | 1 | 4 | 1 | 1 | -1 | -4 | -1 | 1 } |
| {-1 | 1 | 4 | -1 | 1 | 4 | 4 | -4 | 4 | 4 } |
| {1 | 1 | -4 | 1 | -1 | 4 | -4 | -1 | -4 | -1} |
| {-1 | 1 | 1 | 1 | -4 | 4 | 1 | -1 | 4 | 4 } |
| {4 | 4 | -1 | 4 | 4 | -1 | -1 | -4 | 1 | 4 } |
| {4 | -1 | 1 | 1 | -1 | 1 | -1 | -1 | 1 | -4} |
| {-1 | -1 | -4 | 1 | -4 | 4 | -4 | -1 | 1 | 1 } |
| {-4 | -4 | 4 | 4 | 4 | 1 | -4 | -1 | -4 | 4 } |
| {-1 | 4 | -4 | -1 | 1 | -1 | -1 | -4 | 1 | 4 } |
| {-4 | 4 | -1 | 1 | 1 | -1 | -4 | -4 | -1 | 1 } |

Table 4.1

This is the highest capacity network found by delta learning:

| | | | | | | | | | |
|---|---|---|---|---|---|---|---|---|---|
| 0 | 14 | -81 | 142 | 109 | 55 | 91 | 39 | 69 | -95 |
| 229 | 0 | 85 | -261 | -197 | -160 | -108 | -140 | -142 | 227 |
| -93 | -63 | 0 | 54 | 124 | 56 | 97 | -29 | 94 | -75 |
| 81 | -32 | 53 | 0 | -38 | -26 | -80 | -23 | -40 | 67 |
| 75 | -26 | 70 | -70 | 0 | -73 | -34 | -48 | -90 | 99 |
| 145 | 12 | 163 | -124 | -230 | 0 | -100 | -14 | -126 | 190 |
| 122 | 37 | 167 | -163 | -157 | -95 | 0 | -32 | -127 | 121 |
| 266 | -146 | 121 | -190 | -390 | -297 | -89 | 0 | -252 | 271 |
| 133 | 8 | 171 | -116 | -258 | -137 | -107 | -66 | 0 | 154 |
| -85 | 35 | -65 | 71 | 139 | 104 | 42 | 26 | 97 | 0 |

Fig. 4.5 Hopfield network which stores 14 of 15 patterns.

Fourteen of the fifteen patterns are stored by this network.

The capacities shown in these results are significantly superior to the current limits in Hopfield networks. The best values of $t$ all store 6 patterns successfully over 90% of the time for the tested values of $r$ in a neural network with 7 neurons. The successful storage of 14 patterns in a network with only 10 neurons is also a significant improvement over the previous limit. There is also an interesting pattern concerning the largest magnitude in $r$. When the largest magnitude is 3, the least successful storage occurred. The most successful largest magnitude was 4.

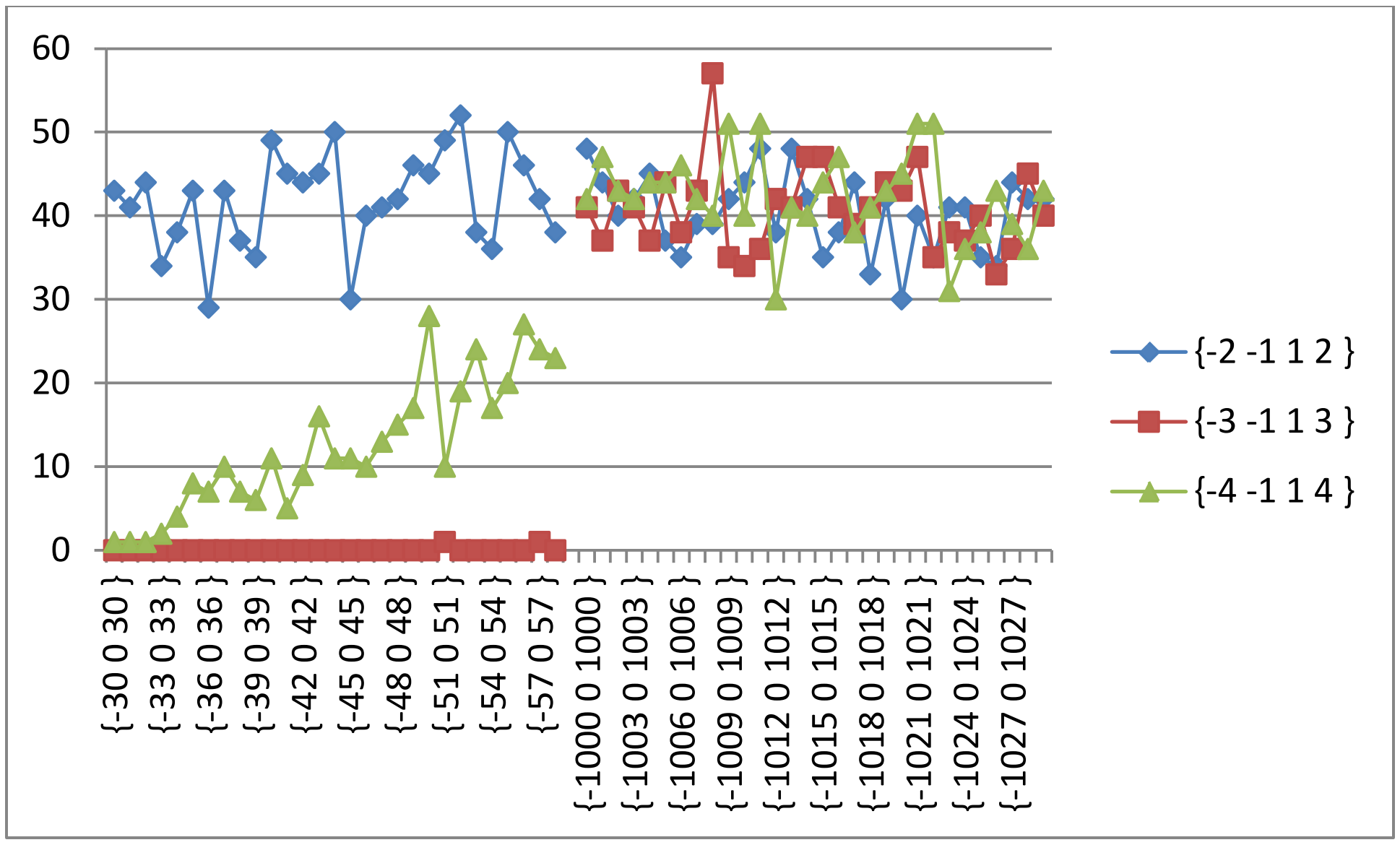

Fig. 4.6 Storing 7 patterns with various thresholds whose distance becomes large.

When attempting to store 7 patterns while mapping to 4 values, the chance of success drops. The following graphs show the percentage of success with various thresholds. The vertical axis is the percentage of storage attempts that successfully stored all 7 patterns. The horizontal axis represents the thresholds used. There is a break between thresholds {-60 0 60} and {-1000 0 1000} because the values in between do not show any unexpected behavior. Once the largest magnitude of the thresholds becomes large enough, the percentage success rate of storing all patterns becomes fairly constant. As shown in the following graph, the point when thresholds become large is dependent on the values being mapped to.

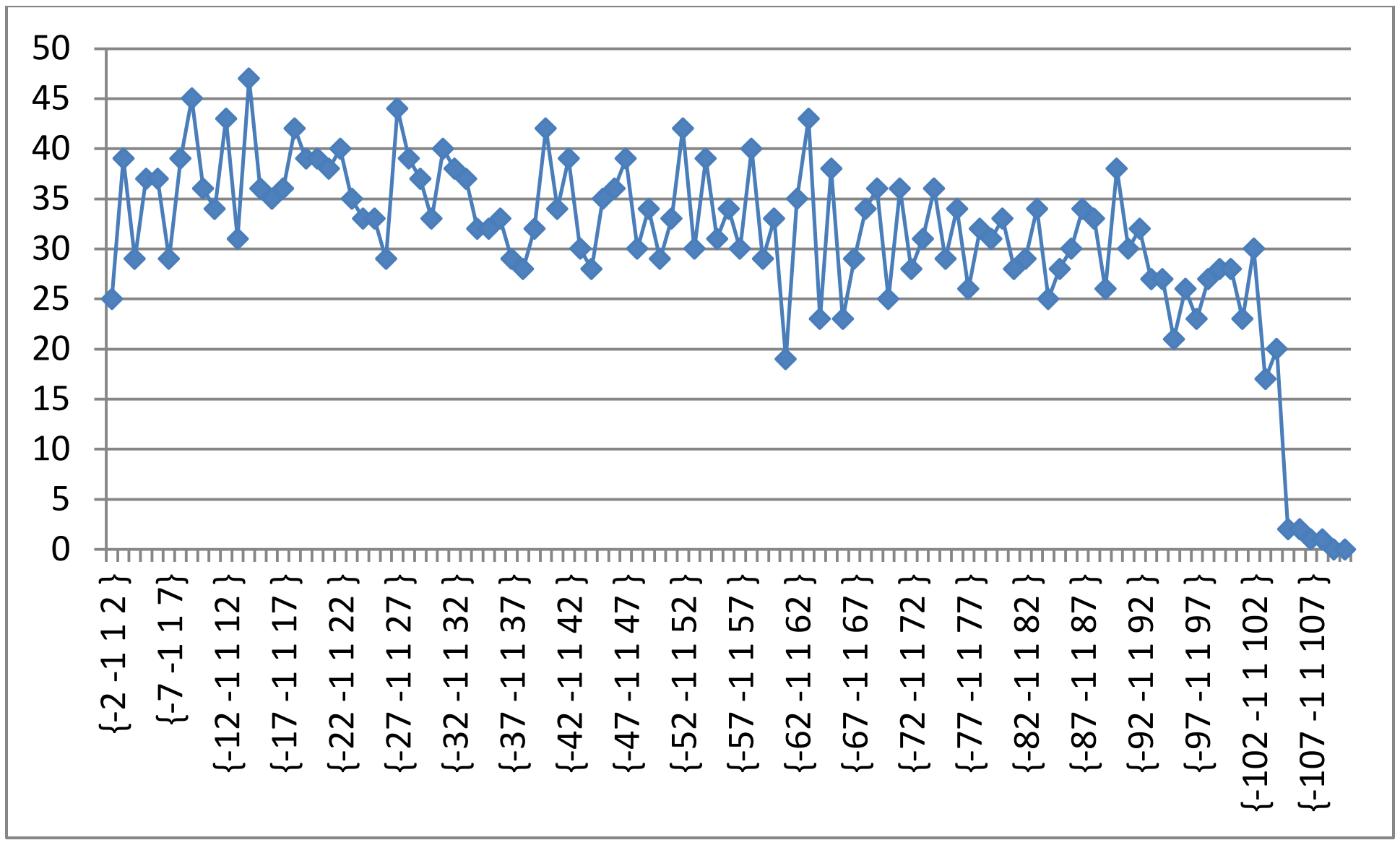

Fig. 4.7 Storing 7 patterns thresholding to values whose distance becomes large.

Figure 4.7 demonstrates the percentage of success for different magnitudes when storing 7 patterns in a network with 7 neurons. Once the magnitude of the mapped to values becomes 109 or larger, the network is unable to store 7 patterns. The thresholds were {-100000 0 100000} for all values mapped to. The vertical axis is again the percentage of success. The horizontal axis represents the values being thresholded to.

Mapping to more than 4 values was also tested. The following graph shows attempts to map to 6 values. Although the success percentage is lower, the information capacity is still much higher for those sets of 7 patterns that do store successfully. The vertical axis is the percentage of successful storage. The horizontal axis represents the different thresholds. The different colored lines represent the different values thresholded to.

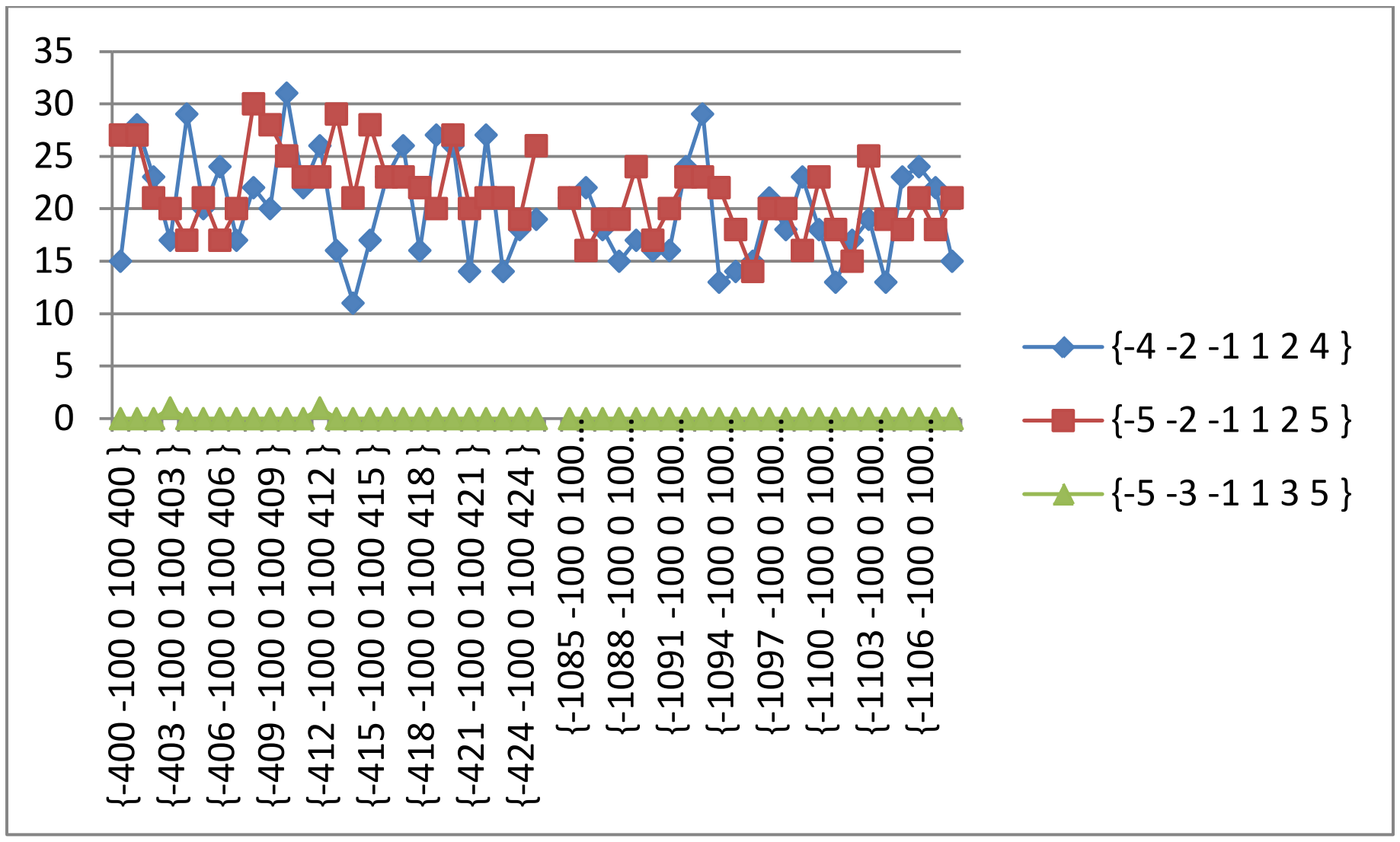

Fig. 4.8 Storing 7 patterns with 6 thresholds.

The following graph begins to reveal the direct relationship between the distances between thresholds and the ability to successfully store all patterns. The axes represent the same values as figure 4.8.

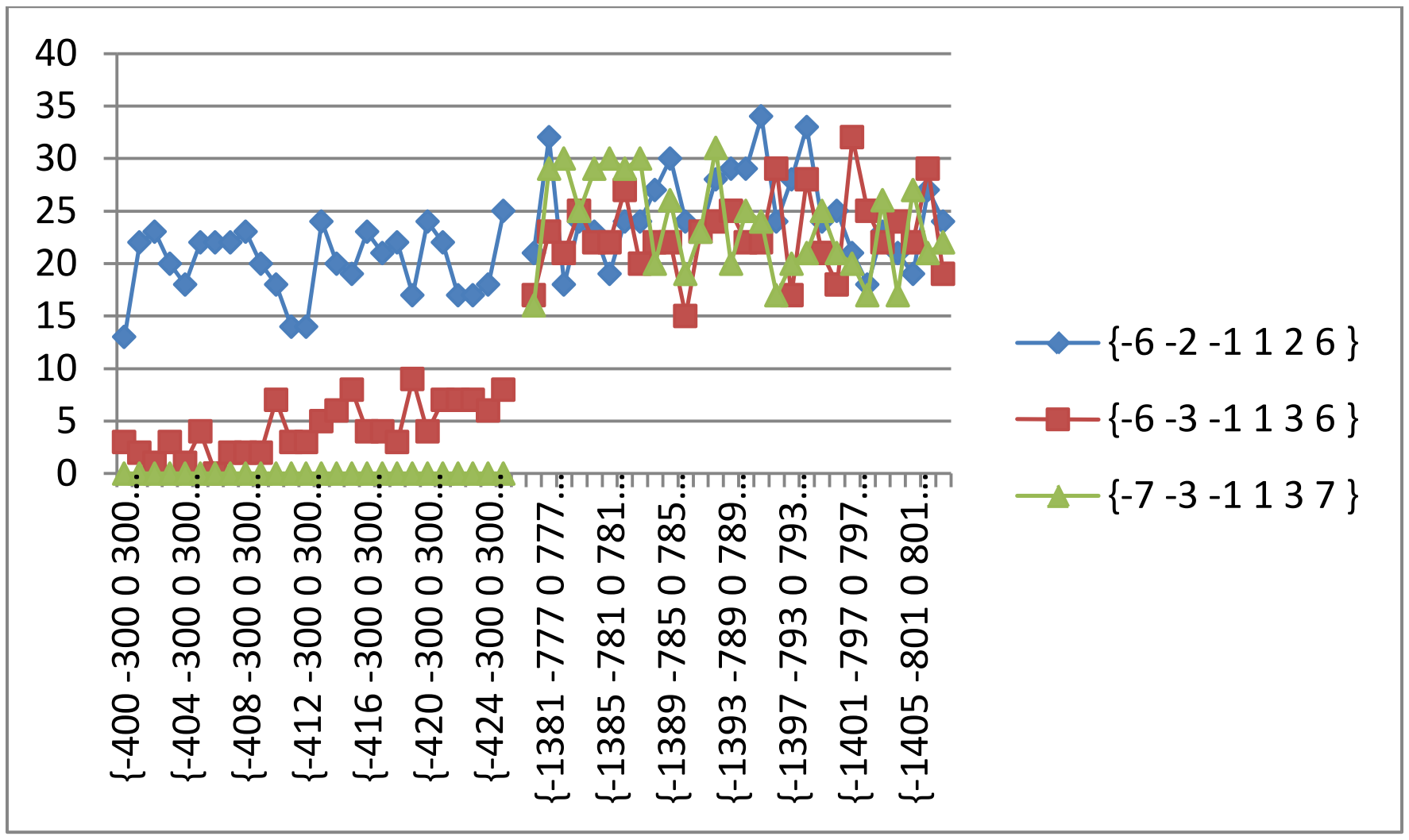

Fig. 4.9 Storing 7 patterns with 6 thresholds with a larger distance between threshold mappings.

Increasing the number of thresholds to map to eight values causes yet another decrease in successful storage percentage, but another exponential leap in terms of information stored. The following graph shows attempts to map to the values {-4 -3 -2 -1 1 2 3 4}. In this graph, we now see a downward trend once the distances between thresholds become too large. The vertical axis is successful storage percentage. The horizontal axis represents the thresholds used.

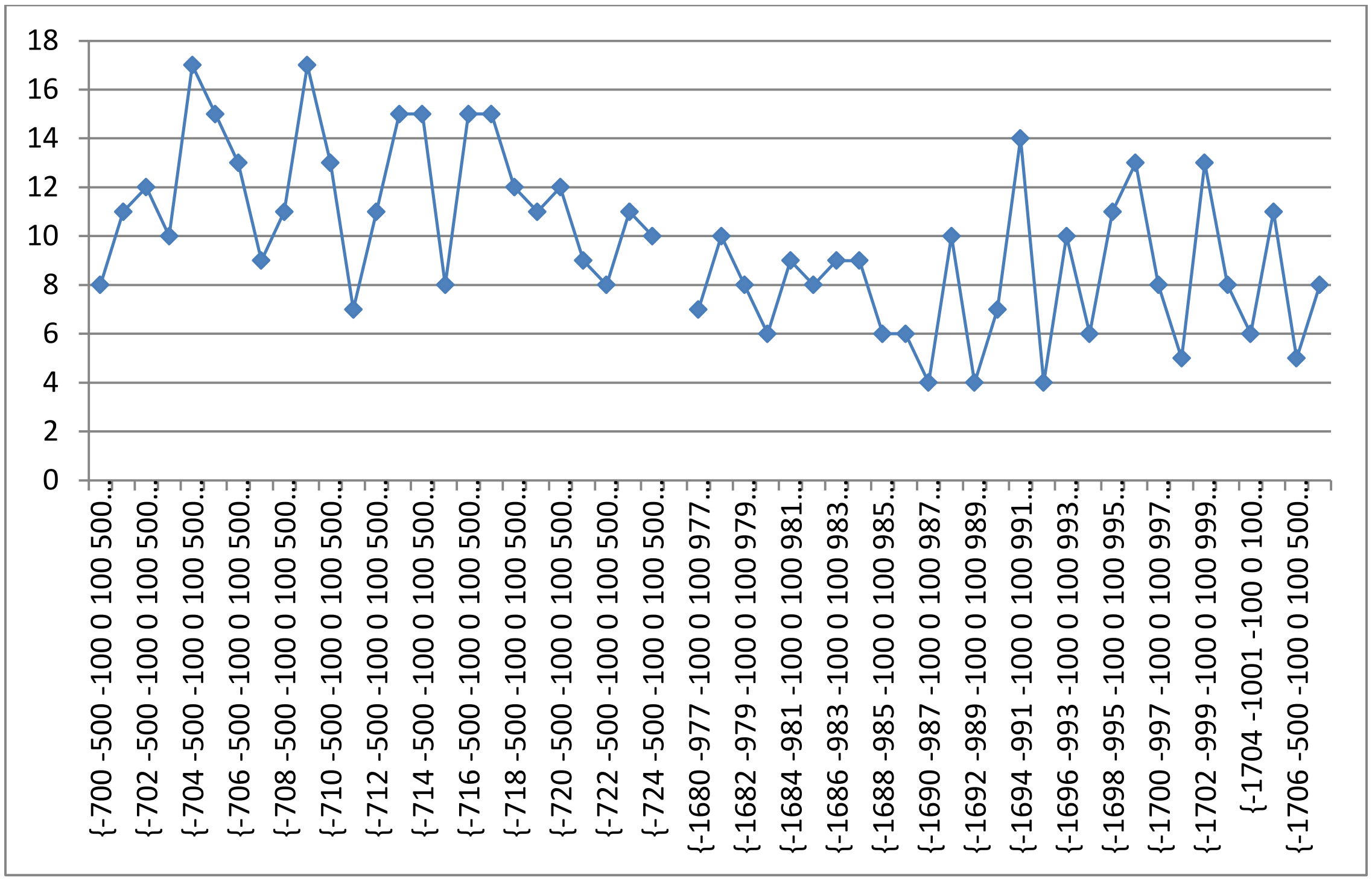

Fig. 4.10 Storing 7 patterns thresholding to 8 values.

As expected, the percentage of successful storage attempts continues to decrease as more thresholds are added. Figure 4.11 shows the relationship between the percentage of groups of patterns successfully stored (vertical axis) and the number of thresholds (horizontal axis).

As shown in the graph, once the number of thresholds increases beyond 40 all 7 patterns will almost always fail to store in a 7 by 7 neural network. It is still possible to store all 7 patterns every once in a while, however. Even mapping to 256 values can successfully store 7 patterns in a 7 by 7 matrix depending on the patterns being stored and the thresholds. The following 7 patterns can be stored with 256 mappings:

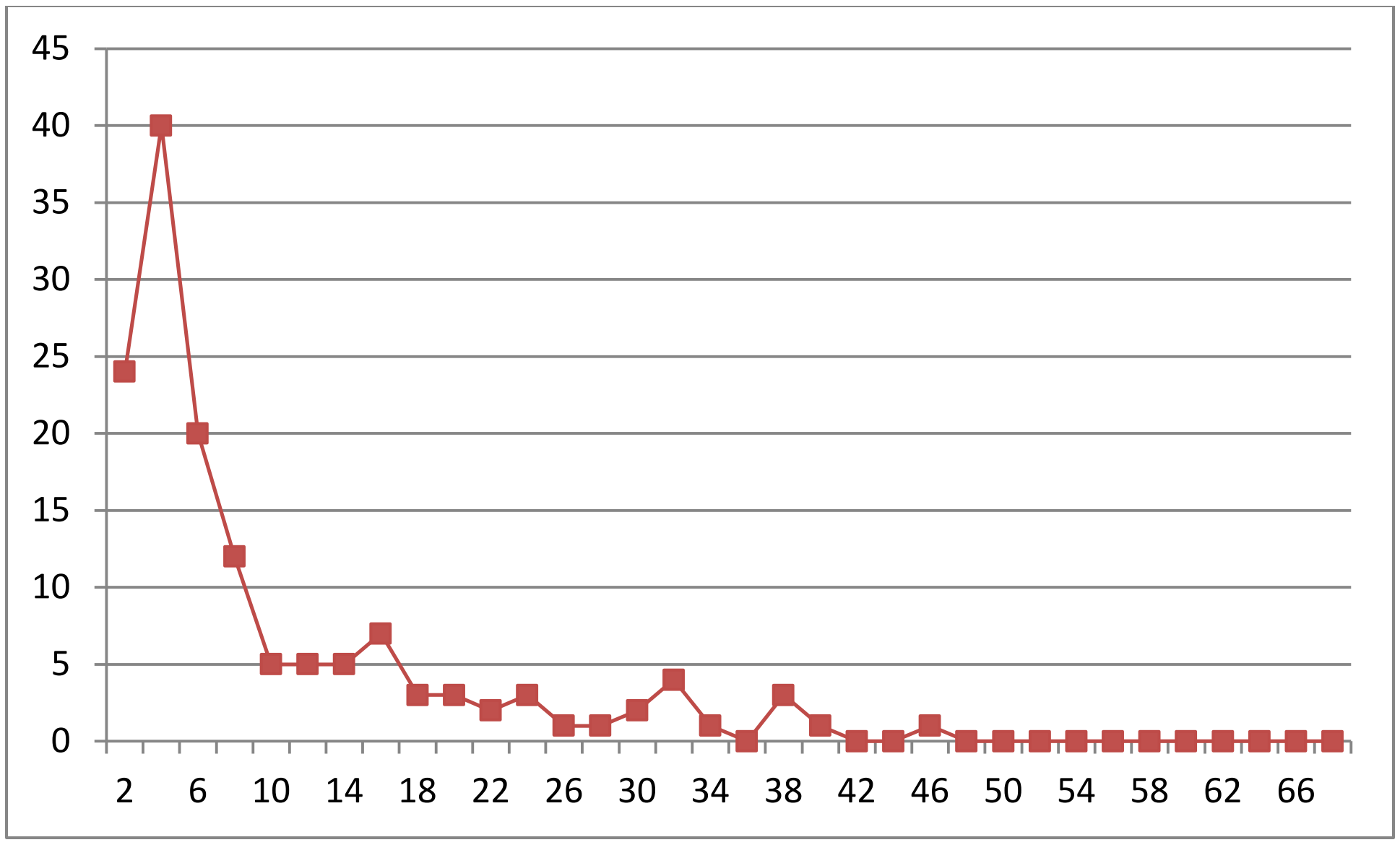

Fig. 4.11 Storing 7 patterns thresholding to increasing numbers of thresholds.

Table 4.2

| | | | | | | |
|---|---|---|---|---|---|---|
| {117 | -120 | -59 | 68 | -117 | 32 | -4} |
| {64 | 76 | -8 | -19 | -17 | -127 | 120} |
| {70 | 30 | 36 | -77 | 62 | 52 | 101} |
| {37 | 84 | -51 | -22 | 67 | 121 | 55} |
| {72 | 96 | 39 | -96 | -32 | -126 | 4} |
| {2 | 6 | -9 | -96 | -38 | 40 | -74} |
| {48 | 100 | 59 | 24 | 8 | -92 | -114} |

The patterns can be stored in the following matrix:

| | | | | | | |
|---|---|---|---|---|---|---|
| 0 | 67720 | 102818 | 37268 | -124156 | 64797 | 42543 |
| 24372 | 0 | -62455 | -22331 | 75055 | -39283 | -25702 |
| 15934 | -26562 | 0 | -14588 | 48708 | -25551 | -16795 |
| 44130 | -74020 | -112215 | 0 | 135362 | -70740 | -46177 |
| -13387 | 22238 | 33518 | 12141 | 0 | 21172 | 13853 |
| 25488 | -42540 | -64939 | -23275 | 78135 | 0 | -26783 |
| 39228 | -65572 | -99100 | -35889 | 119615 | -62520 | 0 |

Fig. 4.12 Hopfield network which stores 7 patterns with 256 threshold levels.

The following values for equation 2.16 allow all of these patterns to be stored by this matrix:

$r$ = {-128, -127, ..., 127, 128}

$t$ = {-5207000, -5166000, ..., -41000, 0, 41000, ..., 5166000, 5207000}

There are several other sets of 7 patterns that can be stored successfully in a 7 by 7 matrix with 256 mappings, but usually only one group of patterns in about 200 can be successfully stored with 256 mappings.

Altering the thresholds while attempting to store the groups of patterns did not result in a very large increase in network capacity. A test was run with {-2 -1 1 2} as $r$ and {-50 0 50} as $t$ for equation 2.16. Without threshold adjustments, the attempt took about 5 minutes and stored 41% of the groups of 7 patterns successfully. With threshold adjustments, the attempt took about 80 hours and stored 42% of pattern groups. This slight increase is clearly not worth the considerable amount of extra time required.

The following graph shows the results when storing a single pattern in a 7 by 7 network with a distance of 10000 between each threshold. The vertical axis is the percentage successfully stored, and the horizontal axis represents the values thresholded to.

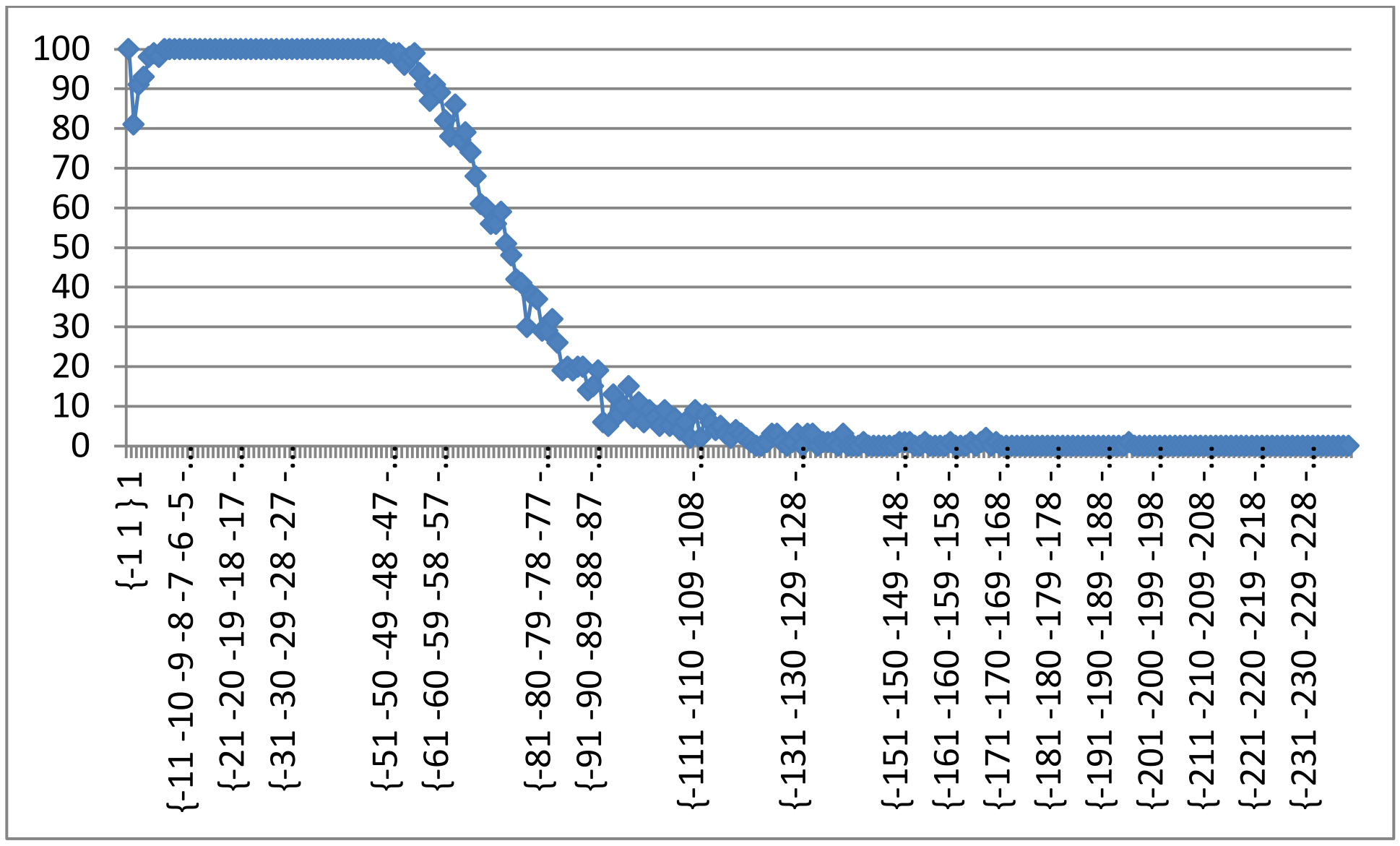

Fig. 4.13 Storing 1 pattern with large threshold counts and 10000 distance between thresholds.

This graph shows that storage of a single pattern can be achieved over 90% of the time with up to 120 thresholds. After that point, the ability to store a pattern drops with each new threshold added. There is also a drop at the very beginning with 4 to 8 thresholds before the general 100% success rate is obtained. The following graph shows the same data except the distance between thresholds has been increased tenfold to 100000.

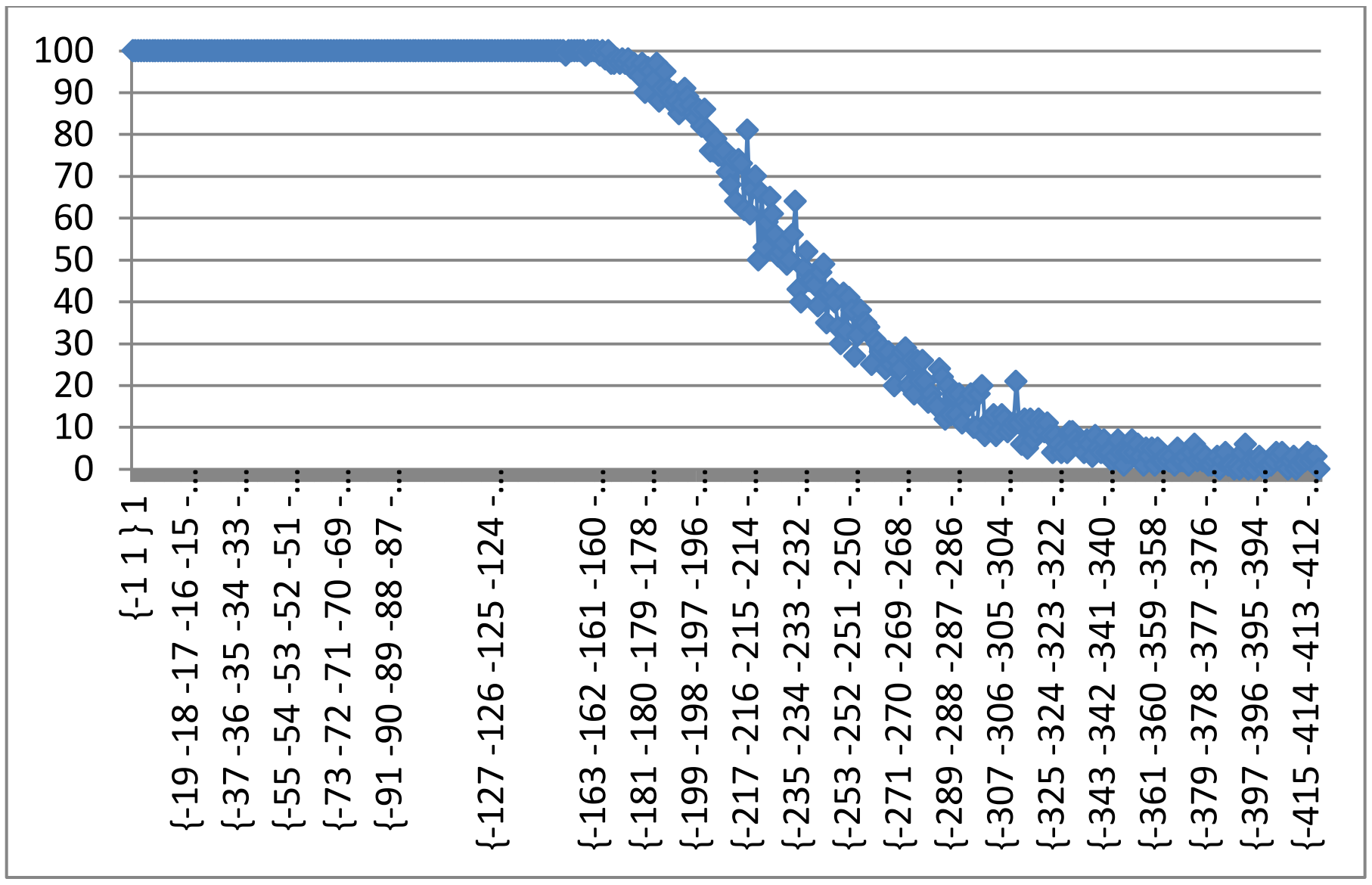

Fig. 4.14 Storing 1 pattern with large threshold counts and 100000 distance between thresholds.

This graph exhibits the same general behavior as the previous one, but there is no drop at the lower threshold counts and the gradual decline begins around 360 thresholds. Increasing the distance between thresholds once again to 1000000 yields the following results.

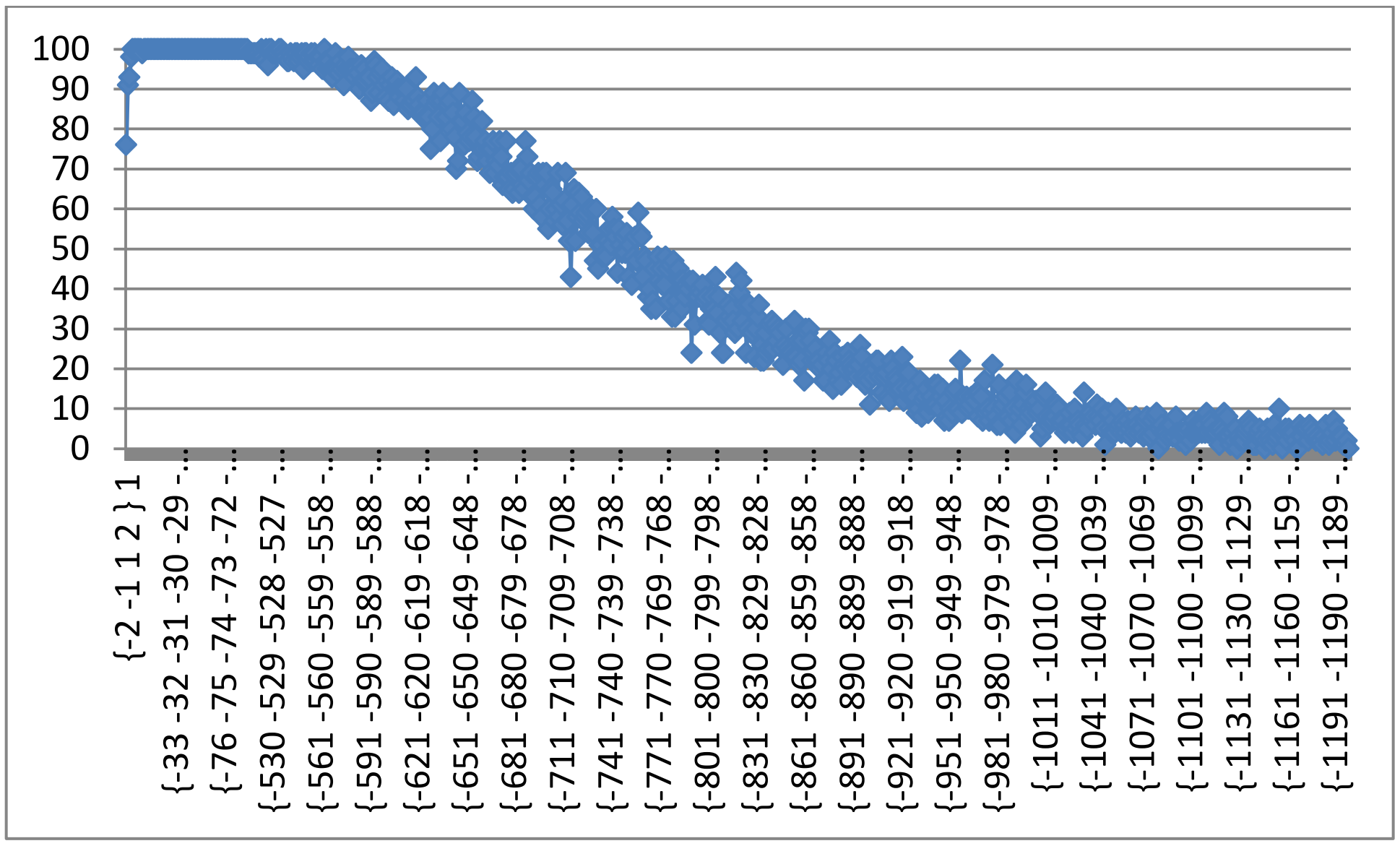

Fig. 4.15 Storing 1 pattern with large threshold counts and 1000000 distance between thresholds.

The results when using threshold distances of one million are similar to the results when using threshold distances of ten thousand. There is again the initial drop for the first 2 to 8 thresholds, and then 90% successful storage is obtained for a large number of thresholds. However, the storage capacity is even greater with one million as the distance between thresholds. The successful storage percentage does not drop below 90% until about 1100 thresholds are used.

These graphs show that the greater the distance between thresholds, the more successful storing a single pattern is with greater thresholds counts.

## V. RECALLING PATTERNS FROM THE STIMULATION OF A SINGLE NEURON

Experiments have shown that thresholding has a negative impact on the ability of a network to recall an entire pattern from the stimulation of a single neuron. To recall a complete pattern from the stimulation of a single neuron one neuron is first set to a value. Then the next neuron is activated with that value, and the activity continues to spread throughout the network until all neurons have returned a value. This process attempts to model similar biological recollections, such as remembering an entire song once the first few notes are heard [37].

The following graph shows the results of several different network sizes. The vertical axis represents the percentage of patterns that are stable patterns in the network that were recalled. The horizontal axis represents the largest magnitude of *r* in equation 2.16. The values of *r* and *t* can be defined as follows:

$r$ = {-X, -X-1, ..., X-1, X}

$t$ = {-10000*(X-1), -10000*(X-2), ..., 0, 10000*(X-2), 10000*(X-1)}

Where X is the value of the horizontal axis. For clarity, when X=1 these values are used:

$r$ = {-1, 1}

$t$ = {0}

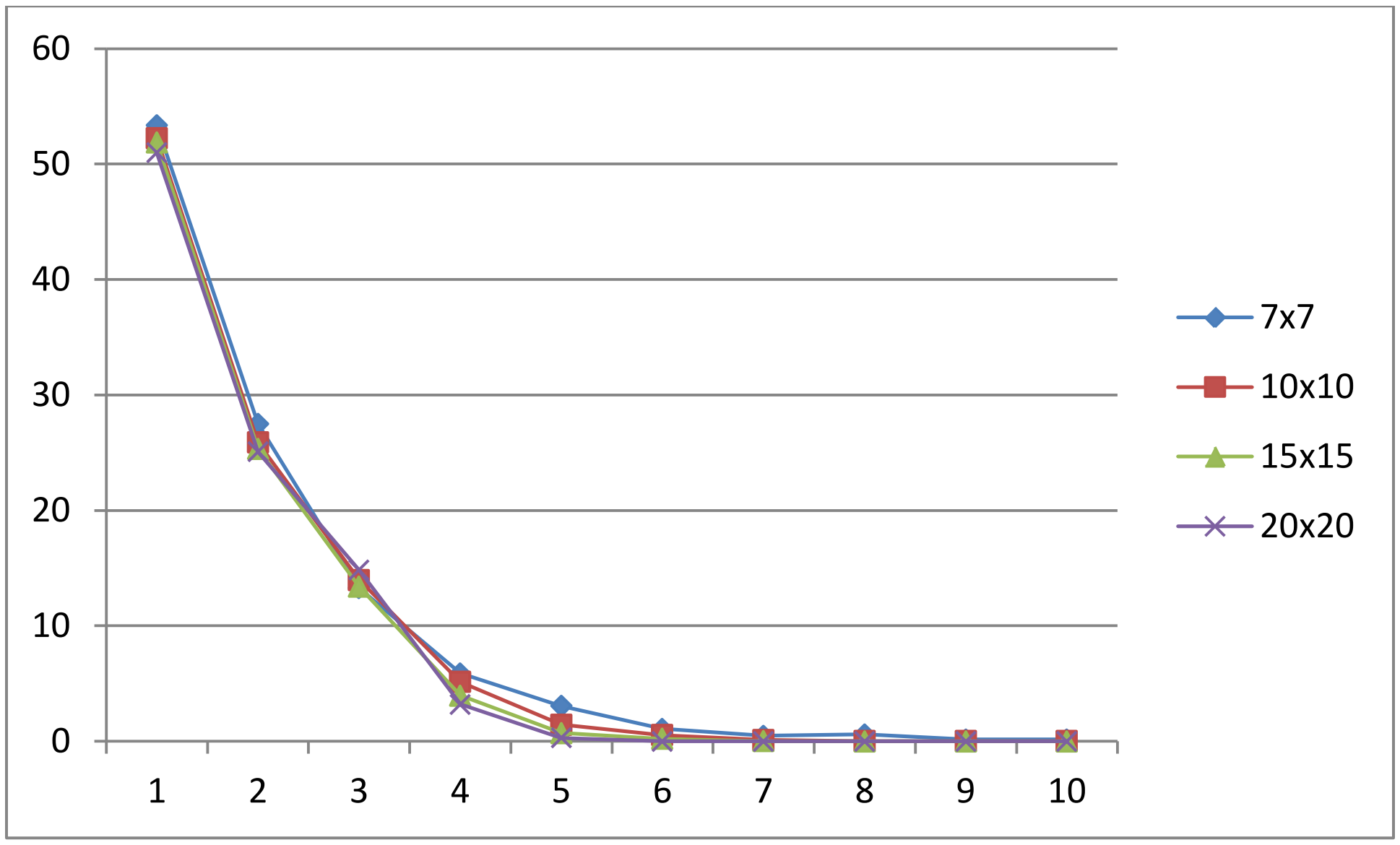

Fig. 5.1 Recalling an entire pattern from the stimulation of a single neuron.

This graph clearly shows that there is only a very slight inverse relation between network size and ability to recall patterns from the stimulation of a single neuron. The more determining factor is obviously the number of thresholds involved in the attempted recall. In regards to recalling patterns from the stimulation of a single neuron, classic binary networks are clearly superior to thresholded networks. This holds true when the thresholds are 10000 apart, the thresholds map to values which are 1 apart, and only setting one neuron to a value and activating allowing the rest to be recalled. For figure 5.1, every possible neuron was set with every threshold mapping to test recall with all possibilities. More than one neuron can be set to a partial stable pattern, but this decreases the useful nature of the recall method.

## VI. EXPERIMENTS REGARDING CAPACITY WITH RESPECT TO NETWORK SIZE

Experiments have shown that the size of the network greatly affects storage capacity. The following graph shows attempts to store *n* patterns, where *n* is the size of the network. The vertical axis is the success percentage, and the horizontal axis is the largest magnitude of the values thresholded to. The distance between thresholds is 10000.

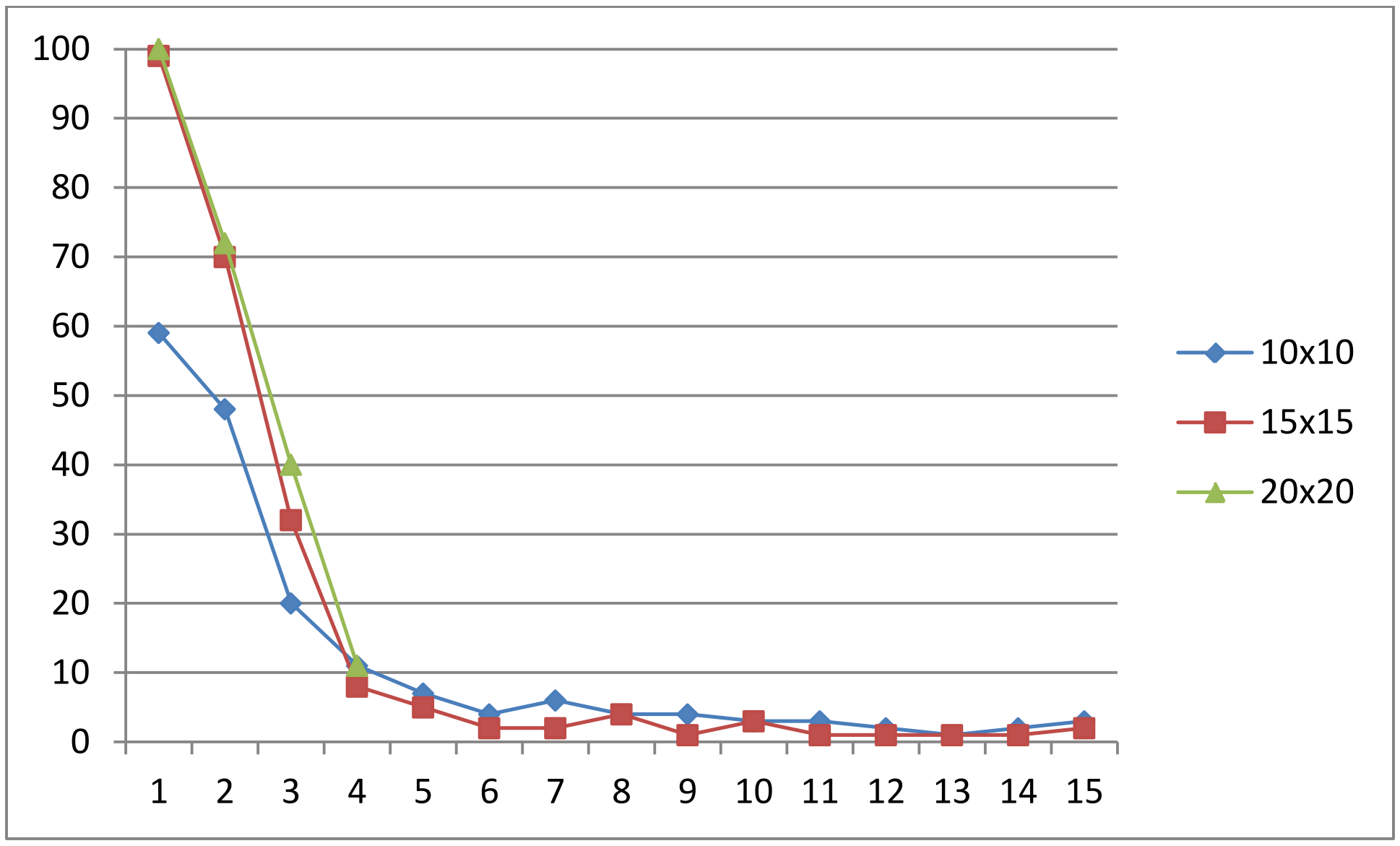

Fig. 6.1 Storing *n* patterns with various threshold counts in an *n* by *n* matrix.

This graph is similar to figure 4.11, and shows a generally direct relationship between size of the network and ability to store patterns. However the size of the network only improves capacity to a point. Once the size of the network becomes large enough, the increased ability to store patterns becomes negligible. The number of thresholds used also plays an important part in how successful the patterns are at storing across all sizes of networks. The more thresholds used, the less successful the pattern storage. However, the more thresholds used, the greater the information capacity in bits per neuron.

The following graph shows the maximum capacity of an *n* by *n* network with respect to bits per pattern. The horizontal axis represents *n*, and the vertical axis is the number of patterns successfully stored divided by *n*. The thresholds were {-100, 0 ,100} and mapped to the values {-2, -1, 1, 2}. Since the number of patterns attempting to store plays a large part in the network's ability to store patterns, an attempt was made by each *n* by *n* network to store varying numbers of patterns from 1 to 2*n*. The largest number of patterns that stored at least 95% of the time was selected for each value of *n*.

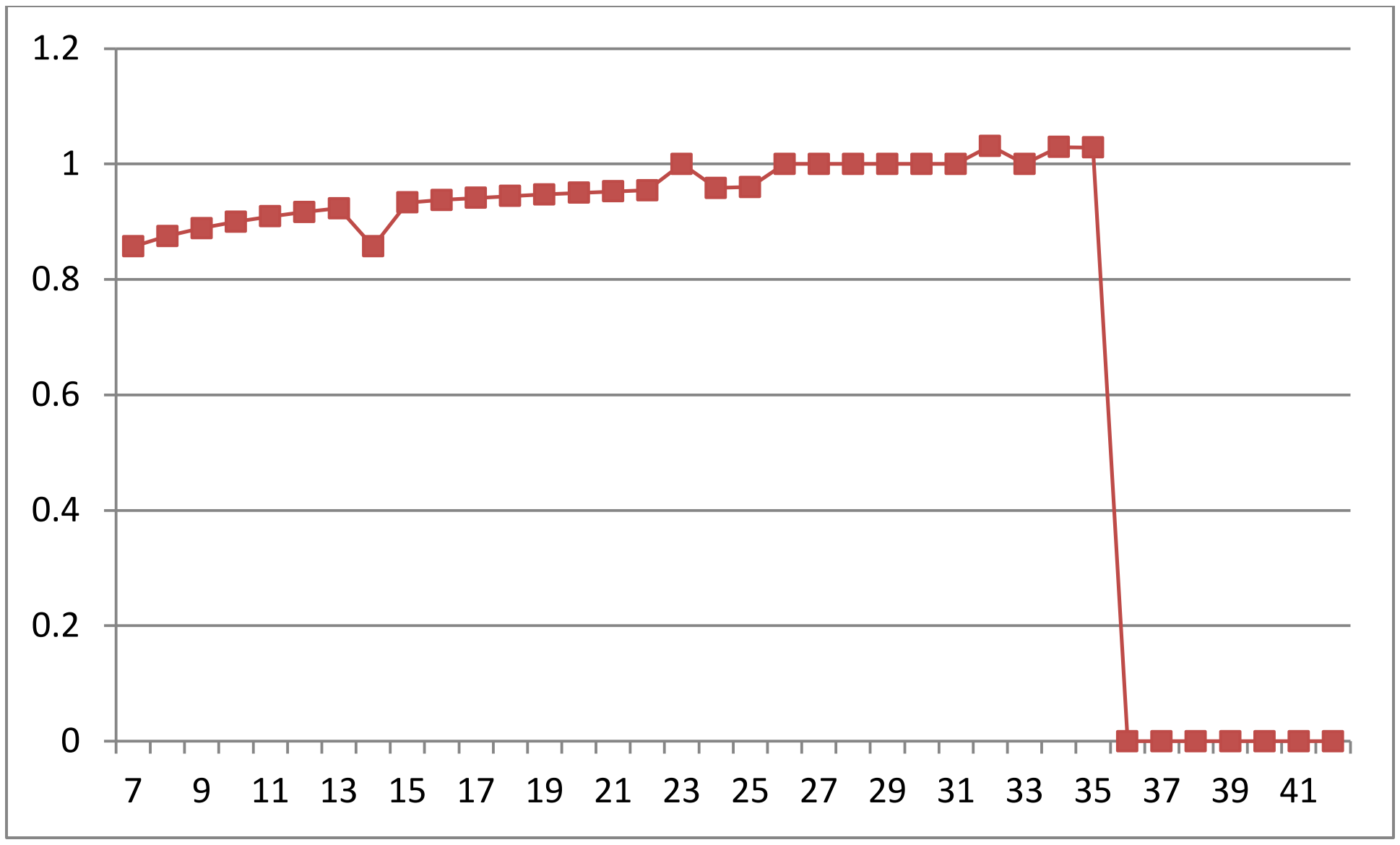

Fig 6.2 Storing the most bits per pattern in *n* by *n* networks.

This graph shows a clear increase in the ability to store groups of patterns until a certain point when the network size becomes too large and even single patterns cannot be stored 95% of the time. This descent is extremely abrupt; there is no gradual decline. The ability to store patterns becomes completely unreliable. The greatest bits per pattern found was in the 35×35 network, where 36 patterns were stored successfully. Since each neuron had 4 possibilities, each neuron represented 2 bits. With 35 neurons that means each pattern stored $4^{35}$ possible values, or 70 bits of information. Since there were 36 patterns stored in the network, 2520 bits, or 315 bytes, can be reliably stored in a 35×35 network with the given thresholds.

## VII CONCLUSIONS

Storage of patterns in a neural network can be improved by repeatedly applying delta learning to an individual memory, the entire group of memories, and various permutations of orders of the entire group of memories. Some limit must be applied to each of these 3 levels of repetition because each technique has the potential to enter into a repeating set of states that will never successfully store all the patterns. When all patterns can be stored successfully, the storage is usually done in a few seconds on a 3 GHz processor.

When mapping to only 4 values, once threshold magnitudes grow sufficiently large the storage percentage remains constant. For instance, when mapping to the values {-2 -1 1 2}, all thresholds greater than {-100 0 100} have roughly between 30% and 50% chance of successfully storing all patterns in a 7×7 network. When mapping to larger values, the distance between successive thresholds plays a critical role in the ability to store patterns. Appropriate distances between thresholds have a direct relationship to the distances between the values being mapped to. If the distances are too small between thresholds, the neural network will not be able to store patterns very successfully. It seems that large distances between thresholds can cause more successful storage, but only up to a point.

It was shown that a 7×7 neural network can map up to 256 unique values and store 7 patterns. However, with each increase in the number of mapped values, the percentage chance of storing 7 patterns successfully drops. Mapping to 4 values has a 40% success chance with appropriate thresholds. Mapping to 6 values only has a 30% chance, and 8 values drops the chance to about 20%. Although the chance of success drops significantly, even at 256 values the chance is not 0, but only less than 1%. The impressive amount of information stored when this is possible makes this case significant. Binary networks can store $2^N$ possible values where $N$ is the number of neurons. Networks with 4 threshold values can store $4^N$ possible values. This leads to the expected formula of $n^N$ possible values where *n* is the number of threshold values in *r*. This means that mapping to 256 values can store over 72 quadrillion values in a single pattern. This has practical application in image recognition networks, which have been monochrome classically. Thresholding to 256 different colors instead of just black and white can help image recognition capabilities greatly.

Storing a single pattern yields a much higher chance of success than attempting to store a group of patterns. Using large threshold values that have a distance between each other of one million and repeatedly applying delta learning, it is possible to store a pattern with up to 1100 thresholded values in a 7×7 network at least 90% of the time. This means that the single pattern stores $1100^7$ possible values or about 70 bits of information. A classic network would require a size of 70 by 70 to achieve this level of storage capacity.

Varying thresholds throughout the storage procedure did not increase storage capacity very much, indicating that aside from greater repetition of delta learning the patterns themselves play an important part in their ability to be stored. It appears that certain groups of patterns do not store as easily as others.

Recalling patterns from the stimulation of a single neuron works best with classic binary networks. As more thresholds are applied, the chance of successfully recalling a stable pattern decreases.

Scaling the methodologies of this paper to larger networks has shown a generally beneficial behavior. The ability to store *n* patterns in an *n* by *n* network grows with respect to *n* but becomes saturated as *n* grows too large. A 10 by 10 network can store 10 patterns much better than a 7 by 7 network can store 7 patterns, but the difference between 15 and 20 for *n* is not very much. This limit could explain certain natural limits observed in biology such as neuron clumps in the brain. The brain is made up of many clumps or groups of neurons instead of just one giant neural network. This behavior of smaller size instead of growing out of control can also be seen on a larger scale in the size of animals. The more advanced animals such as monkey, humans, and dolphins are all medium sized compared to the rest of the animal kingdom. Creatures that are very small are too simple to be particularly complex, and animals that are too large must spend resources retaining that size. Human languages also exhibit this behavior of grouping. Small words are often confused with one another, but words with about 7 letters are easily distinguishable. Words that are longer than that can become confused. The delta learning method applications with thresholds used in this paper have captured some of these aspects of biological neural networks using artificial Hopfield networks.

## ACKNOWLEDGEMENTS

The authors would like to acknowledge the Oklahoma State University High Performance Computing Center for the use of the Cowboy supercomputer in this research.